\documentclass[sigconf]{acmart}

\usepackage{subfig, float}
\usepackage{amssymb}
\usepackage{amsmath}
\usepackage{graphicx}
\usepackage{array}
\usepackage{balance}
\usepackage{colortbl}

\def\P{\mathbb{P}}

\AtBeginDocument{%
  \providecommand\BibTeX{{%
    \normalfont B\kern-0.5em{\scshape i\kern-0.25em b}\kern-0.8em\TeX}}}

\setcopyright{rightsretained}
\copyrightyear{2019}
\acmYear{2019}
\acmConference[AdKDD '19]{AdKDD}{August 05, 2019}{Anchorage, Alaska}
\acmBooktitle{AdKDD '19, August 05, 2019}
\acmYear{2019}
\acmMonth{8}
\acmPrice{15.00}
\begin{document}

\title{Time-Aware Prospective Modeling of Users for Online Display Advertising}

\author{Djordje Gligorijevic}
\email{djordje@verizonmedia.com}
\affiliation{
  \institution{Yahoo Research}
  \streetaddress{701 First Avenue}
  \city{Sunnyvale}
  \state{CA}
  \postcode{94089}
}

\author{Jelena Gligorijevic}
\email{jelenas@verizonmedia.com}
\affiliation{
  \institution{Yahoo Research}
  \streetaddress{701 First Avenue}
  \city{Sunnyvale}
  \state{CA}
  \postcode{94089}
}
\author{Aaron Flores}
\email{aaron.flores@verizonmedia.com}

\affiliation{
  \institution{Yahoo Research}
  \streetaddress{701 First Avenue}
  \city{Sunnyvale}
  \state{CA}
  \postcode{94089}
}

% \affiliation{%

% \affiliation{%

% \affiliation{%

% \affiliation{%

% \affiliation{%



\renewcommand{\shortauthors}{Gligorijevic Dj., et al.}

\begin{abstract}
Prospective display advertising poses a great challenge for large advertising platforms as the strongest predictive signals of users are not eligible to be used in the conversion prediction systems. To that end efforts are made to collect as much information as possible about each user from various data sources and to design powerful models that can capture weaker signals ultimately obtaining good quality of conversion prediction probability estimates. In this study we propose a novel time-aware approach to model heterogeneous sequences of users' activities and capture implicit signals of users' conversion intents. On two real-world datasets we show that our approach outperforms other, previously proposed approaches, while providing interpretability of signal impact to conversion probability.
\end{abstract}





\maketitle

\section{Introduction}
Online Display advertising has been one of the fastest growing industries in the world. In the U.S. alone, this industry amassed \$100 billion dollars in 2018\footnote{https://www.iab.com/wp-content/uploads/2019/05/Full-Year-2018-IAB-Internet-Advertising-Revenue-Report.pdf}. The concept of online display advertising (DA) is developed with the purpose of showing the most relevant ads to users anywhere online. The DA industry is composed of three major components: Supply Side Platforms (SSPs) who realize ad display opportunities on registered websites with user traffic and send ad requests to the following component, the online ad exchanges, who organize online auctions and forward the ad calls to several Demand Side Platforms (DSPs), a third component of the system, to bid on them. In order to have their ads shown to users, advertisers rely on DSPs to reach relevant users through ad display opportunities, bid on the auctions and display advertisers' ads. It is the job of the DSPs to learn which users could be interested in the advertisers' products and could become their business in the near future. In order to achieve that, DSPs try to learn as much as possible about users, by collecting their online footprints through the data collected from advertisers websites, won auctions, third-party data providers and from owned-and-operated (O\&O) properties.

Much of the DAs' business historically has been \textit{retargeting}, a special case where ads are displayed to the users who have already shown interest in advertisers business. The goal of retargeting is to periodically remind users of the advertisers' products and hopefully generate conversions. However, as this particular form of DA is unlikely to bring \textit{new} customers to the advertisers, they have shown increased interest in \textit{prospective} targeting of users. The goal of \textit{prospective} targeting is opposite of retargeting -- users who have shown interest into advertisers business in the recent past should be excluded, and the goal becomes to generate new users as both visitors and converters for the advertiser. While the definition of retargeting users may significantly vary from one advertiser to another, in terms of the general advertising funnel (stages in which users are placed with respect to their probability of purchases of advertiser products \cite{geminix_kdd}), prospective targeting should focus on users who are in the upper funnels (users further away from the conversion funnel). Conversely, in terms of advertising funnel, retargeting focuses on users in the very low funnel stages (users very close to conversion). 

Prospective modeling of users poses a particularly difficult task for DSPs, as the direct signals of users interests (such are visits to advertisers website or recent conversions with the same advertiser) are no longer viable to be used. To maintain the high performance of user modeling, DSP is given a challenging task to generate powerful models which are able to detect relevant, yet weaker, signals users leave in their online trails and use them to the fullest extent. An example of such signals could be users' recent wedding related invoice which could signal potential interest in user purchasing furniture or flight ticket to the honeymoon, whereas any signals related to furniture or flight browsing on advertiser's website could not be consumed.

Moreover, a very important aspect of prospecting user modeling is explainability. Advertisers often require DSPs to provide insights into how predictions were made, what individual signals and what signal combinations seemed important during the modeling process. For the case of prospective modeling, these signals when interpreted can bring exceptional value to the advertiser, as they would be able to fine tailor future campaigns for different user groups that resonate better and reach more consumers which they potentially couldn't before.

\begin{figure}[t!]
	\centering
	\includegraphics[width=0.49\textwidth]{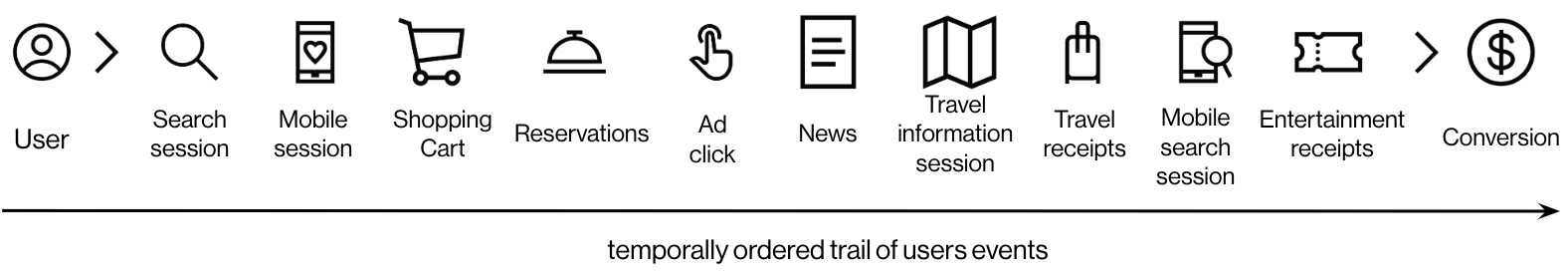}
	\caption{Visualization of user activity sequence with different groups of activities ordered by the time they occurred and ending with the action of advertiser's interest.}
	\label{fig:user_activity_trail}
\end{figure}

% Generating features from non-retargeting (prospecting) signals can be a tedious task due to different definitions of time aspect of retargeting signals with different advertisers and also due to the heterogeneity of users signals where retargeting signals can be intertwined with non-retargeting signals. 
To create a generic view on signals users leave, the most natural choice is to create a time-ordered sequence of activities user performed collected by the DSP. An example of one such sequence is provided in the Figure~\ref{fig:user_activity_trail} where we observe multiple interactions of the user with different online properties such as mobile and desktop search, email receipts, reading news and interacting with ads. 
% Unlike traditional feature handcrafting, 
These trails of user's activities provide insight into the sequence of actions rather than sequence-oblivious features, moreover, actions always have assigned timestamp which carries a significant amount of additional information in terms of how close the subsequent events were or how much time passed between activity and event of interest (i.e. conversion). 
% However, generating data in this manner means that non-relevant activities could be included in these sequences, which poses an additional challenge for the models to learn to ignore them while being able to select the important signals.
Modeling sequences of user events have been proposed in the past with great success \cite{gligorijevic2018sigir,geminix_kdd}, however, to the best of our knowledge, never for prospective modeling of users. Moreover, utilizing activities data to the full extent such as temporal aspect has been largely ignored when modeling conversions in DA. 

We summarize the contributions of this work below:
\begin{itemize}
    \item We motivate and propose the problem of prospective targeting in display advertising. To the best of our knowledge we are the first to discuss research for prospective modeling of users' interests.
    \item We propose sequence learning approach to model time-ordered heterogeneous user activities coming from multiple data sources.
    \item We propose a novel time-aware mechanism to capture temporal aspect of events and thus capturing better their relevance to the conversion.
\end{itemize}

% The rest of the paper is organized as follows. In Section~\ref{sec:background} we describe previous approaches to model conversions in DA and provide motivation behind our choices of models. Problem set-up is then provided in Section~\ref{sec:problem_setup} and methodology is described in Section~\ref{sec:methodoogy}. Datasets used to evaluate the methods baselines are described in Section~\ref{sec:data}. Experimental results are discussed in Section~\ref{sec:experiments} and finally study is concluded and future work is described in Section~\ref{sec:conclusions}.

\section{Background and Related work}
\label{sec:background}
A brief overview of online advertising is given to stress the importance of predicting future conversions and to place it in the grander ecosystem. Additionally, relevant prior works on conversion prediction will be mentioned and their contributions discusses with respect to this study.

\subsection{Online Advertising}
Major DSP platforms for display advertising (e.g., Google DoubleClick, Verizon Media DSP) allow advertisers to sign up and run campaigns and lines. Every advertiser can create multiple campaigns and multiple lines within each campaign that target certain activity. Activities, for example, can be ad clicks or conversion activities (definition of which varies from one advertiser to another). The task for DSP platforms becomes to run advertiser's lines and serve users such that key performance indicators (KPIs) goals are reached. This is achieved by participating in online auctions for different ad opportunities. An important aspect of participating on online auctions is deciding on the value of the ad opportunity as the maximum bid. For conversion predictions the maximum bid is often controlled by the probability of user converting not long after ad is displayed, or more precisely, the maximum bid is defined as a factored conversion probability:
\begin{equation}
maximum\_bid = \alpha * pCVR.
\end{equation}
Thus, the estimate of conversion probability $pCVR$ is one of the key components in the DSP business that drives performance and directs the system towards displaying ads to relevant users. Similar relation can be used for click prediction lines.

\subsection{Modeling users' conversion prediction}
In large scale advertising setups, conversion probabilty estimation has been succesfully tackled throught logistic regression models \cite{bhamidipati2017cikm}. However, manually  designing  and  selecting  features  requires  substantial investment  of  human  time  and  effort,  and  utility  of  such  generated  features  is largely dependent on the domain knowledge of human experts curating the features. Moreover, since typical applications are nonlinear, considering feature interactions (e.g cross-features) quickly becomes prohibitively expensive due to a
combinatorial explosion \cite{mcmahan2013ad}. 

Recently, representation powerful deep learning models have also been proposed for CTR and CVR prediction, e.g., factorization machines \cite{pan2019predicting} for CVR or deep residual networks \cite{shan2016deep} for CTR that tackle problems of learning non-linear interactions of features. 
Also, models that capture inforamtion from the sequence such are RNNs have been proposed recently \cite{cui2018modelling,zhang2014sequential,arava2018deep,gligorijevic2018sigir} and the reportedly perform significantly better than their non-sequential counterparts.
Moreover \cite{arava2018deep} and \cite{geminix_kdd} have used sequences of events from heterogeneous data sources, while \cite{arava2018deep} has additionally proposed adding temporal information of events as an additional source of information to better model sequence for conversion attribution task.

It is worthwhile noting that there are currently no notable papers describing the use case of prospective user conversion modeling.

\section{Methodology} 
\label{sec:methodoogy}
% We provide formulation of our prediction task, followed by describing the proposed model designed for tackling the formulated problem. Finally interpretability of the models predictions is discussed.
In this section we discuss proposed mode and its interpretability.

% \subsection{Problem set-up} 
% \label{sec:problem_setup}
% We are given a trail users online activities $\mathcal{A} = \{a_i | i = 1 \ldots M\}$.  Each activity $a_i$ is represented as a tuple of the string representation of the activity and corresponding timestamp in milliseconds when it occurred $< \textnormal{act\_name}, \textnormal{timestamp}>$. Some activities can be labeled as retargeting activities or conversions as specified by the advertiser, and rules are defined over activity properties. 
% Once all user activities are ordered by time, retargeting events are removed and the trail is cut a day before conversion happened, or no cutting is performed on the trail if there was no conversion activity. After the cut, if any, conversions are collected and used as a target for supervised learning. Learning task of the algorithms is that given a user activity trail (processed for prospecting modeling) predict if the user will convert for the advertiser in the near future.

\subsection{Proposed Approach}
\label{sec:proposed_approach}
\begin{figure}[ht!]
	\centering
	\includegraphics[width=0.4\textwidth]{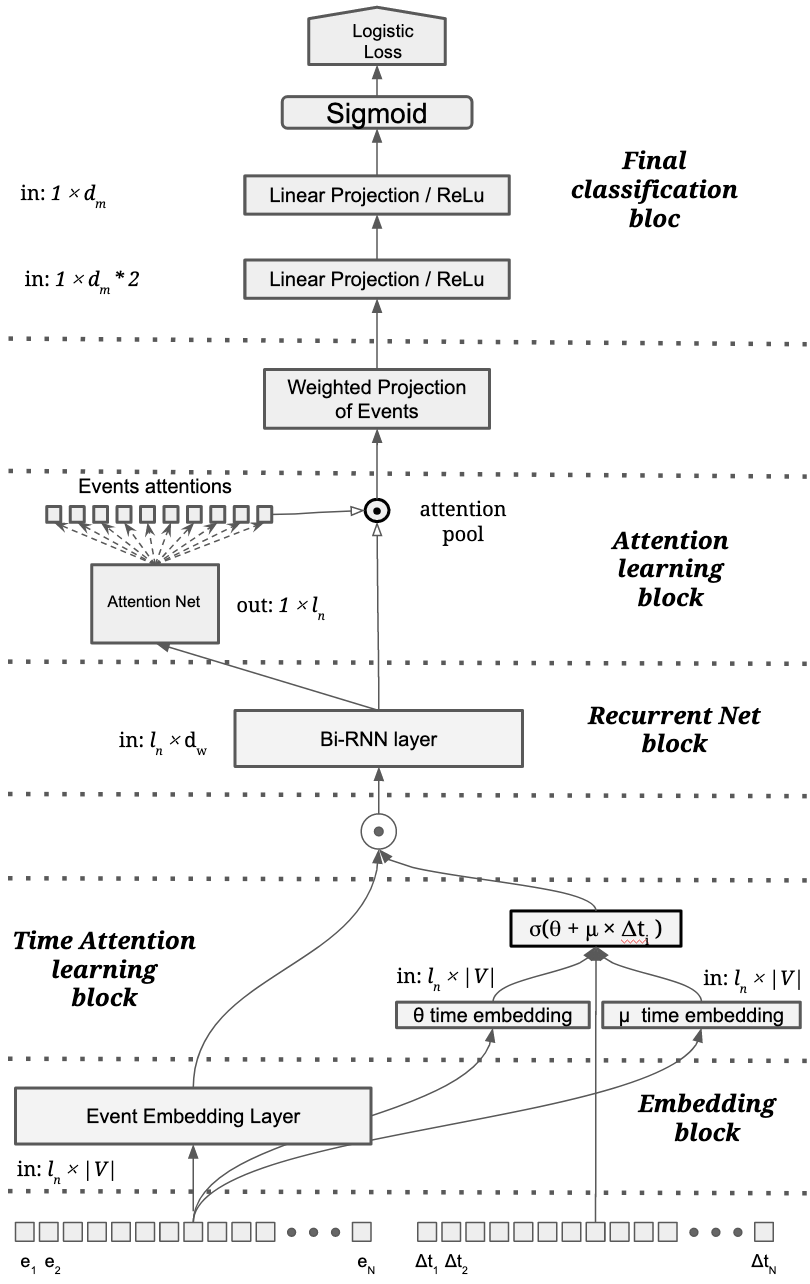}
	\caption{Graphical representation of the DTAIN model}
	\label{fig:model}
\end{figure}

We propose a novel model - Deep Time Aware conversIoN (DTAIN) model (Fig.~\ref{fig:model}).

The DTAIN model takes sequence of events $\{e_i | i = 1 \ldots N\}$ and time difference of events' timestamps and the time point of prediction (usually timestamp of last event in a sequence) as inputs. It then forwards this information through 5 blocks specifically designed for this task to learn conversion rate prediction.
% learns three separate embeddings, one for events themselves and two more single-dimensional embeddings for modeling temporal information. Through a series of layers, including bi-direction LSTM and attention layers embedding for the user is obtained, and then passed through a series of linear projections with non-linearities to learn conversion rate prediction.

\subsubsection{Blocks of the DTAIN model}

\paragraph{Events and Temporal information embedding}
Embeddings of events and temporal information are performed in two separate parts of the network. First, $l_n$ events in the user's trail and $l_n$ timestep information associated to them are embedded into vectors $h_{e_i}$ of $h_{e_i} \in \mathbb{R} ^ {d_{w} = 300}$ dimensional common space (\textit{Embedding block}).

\paragraph{Temporal attention learning} Each event $e_i$ is also associated with two additional single-dimensional learnable parameters as $\mu_{e_i}$ and $\theta_{e_i}$ of $\mu_{e_i}, \theta_{e_i} \in \mathbb{R} ^ {d_{t} = 1}$. These parameters are designed to model the temporal increment $\Delta_t$ as time difference between current state $i$ and the state of interest $j$ (i.e. timestep when pCVR is served):
\begin{equation}
\Delta_t = \tau_{e_j} - \tau_{e_i}
\end{equation}
\begin{equation}
\delta(e_i, \Delta_t) = S(\theta_{e_i} - \mu_{e_i} \Delta_t)
\end{equation}
\begin{equation}
S(x) = \frac{1}{1 + e^{-x}}
\end{equation}

\noindent $\delta(e_i, \Delta_t)$ captures the influence of the current event to conversion with $\theta_{e_i}$ measuring initial influence and $\mu_{e_i}$ measuring the change of the influence of the event with the time difference. Smaller $|\mu_{e_i}|$ refers to events whose influence does not change as we observe the event through different points in the users trails, while larger $|\mu_{e_i}|$ means that position and time of the event is very important for measuring its effect on conversion probability (given that the $\Delta_t$ is always positive and provided that $\theta_{e_i}$ doesn't change, larger positive values of $\mu_{e_i}$ would mean that the temporal scores is closer to $0$, and larger negative values that is closer to $1$).
Similar ways of modeling temporal increments can be seen in known results of Euler's forward method \cite{cao2018learning} for modeling change of state in dynamic linear systems. In our case, we opted for using time information as an event-level contribution to the final task, thus Sigmoid function was used to transform $\theta_{e_i} - \mu_{e_i} \Delta_t$ into probability between 0 and 1. Rather than choosing Softmax layer which would force total influence of all events to be equal to one, we opted in to use Sigmoid to model influence of each event individually given their own specific influence factors.
This approach allows us to model same events that happened multiple times withing same user trail differently, i.e. giving more attention to events that happened more recently. 

Other formulations of time information were given in \cite{arava2018deep}, however their approach only includes the cases of strict time decay effect where only events which happened close to prediction time may pass full information through the classifier. Similarly to \cite{bai2018interpretable} we learn event-specific initial and time influence factors which we use to control how much information passes from each event embedding into the first non-linear layer of the model.

The learned embeddings and contributions of each event are then summarized to obtain new event representation $v_{e_i}$:
\begin{equation}
\forall_{h_{e_i} \in \{i = 1 \ldots N\}} \forall_{\delta(e_i, \Delta_t) \in \{i = 1 \ldots N\}}  v_{e_i} = h_{e_i} * \delta(e_i, \Delta_t)
\end{equation}
Resulting again in $v_{e_i} \in \mathbb{R} ^ {d_{w} = 300}$ dimensional space. This way of modeling allows for model interpretability, as for each event we can measure its initial and time influence factor and interpret their values as described above.

\paragraph{Recurrent Net block}
The resulting embeddings of events are then fed into bi-directional RNN model (with GRU cells used for both forward and backward pass networks), our first non linear layer in the model:
\begin{equation}
    g_{e_1}, g_{e_2}, \ldots, g_{e_N} = biRNN(v_{e_1}, v_{e_2}, \ldots, v_{e_N}, \theta_{GRU})
\end{equation}
Bi-directional RNN's ensure that the model learns complex relations between events, which is in particular important for user trails where evens may be grouped by sessions which carry higher order information than the events themselves \cite{Gligorijevic2019}. The resulting embeddings $g_{e_i}$ are embedded into  $v_{e_i} \in \mathbb{R} ^ {d_{m} = 200}$ dimensional space.

\paragraph{Attention learning block}
In order to learn rich representations of user's trail, it is imperative to focus on events that carry the most information. To learn representations that focus on important parts of the user trail we employ a dedicated attention mechanism on top of sequence modeling features \cite{Gligorijevic2019}.
Employed attention block yields event scores, that highlight events of greater importance for the task at hand. In out particular case, attention model is implemented as two-layered individual neural network $s_q(v_e; \theta_e)$with Softmax at it's final layer:
{\begin{equation}
	t_{e_i} = \frac{\exp(s_e(g_{e_i}; \theta_e))}{\sum_{i = 1}^{l_n} \exp(s_e(v_{e_i}; \theta_e))}.
\end{equation}}
Neural networks $s_e(v_{e_i}; \theta_e)$ learns real valued scores for each $i^{th}$ event in a given user trail. Attentions learning in the DTAIN model is coupled with the entire network (end-to-end).

Event attentions $t_{e_i}$ are then used to re-weight their input representations $g_{e_i}$ and to obtain compact representation of the entire sequence $s = \sum_{i}t_{e_i}*g_{e_i}$.
There are other ways of obtaining compact representations $s$, such as sum, average or max of individual event vectors. However, our experiments, as well as available literature \cite{zhai2016deepintent,gligorijevic2018sdm}, demonstrate that such strategies are inferior to using attention.

\paragraph{Learning to predict from the resulting representation}
The summarized user trail representation from previous block is finally fed to a sequence of fully connected layers with ReLU nonlinearities before finally passing through a sigmoid layer $\sigma(\cdot)$ to obtain the probability of conversion (pCVR).

Finally, to optimize the parameters of DSM (denoted as $W$ in remainder of the text), we have obtained logistic loss $\mathcal{P}$ for the CTR prediction based on logits from the topmost layer:

\begin{equation}
\mathcal{P}(W) = -\frac{1}{N} \sum_{n=1}^{N}(y_n \log(\hat{y}_n) + (1-y_n)\log(1-\hat{y}_n) ),
\label{eq:logistic_loss}
\end{equation}
where $\hat{y}_n$ are obtained logits after final sigmoid layers and $y_n$ is conversion label for the $n^{th}$ user trail. 

Weights are initialized by a truncated normal initializer. To optimize $\mathcal{L}$, we use Adam~\cite{kingma2014adam} with a decaying gradient step.

\section{Data Description}
\label{sec:data}

\paragraph{RecSys 2015 challenge} 
We conducted conversion prediction experiments on publicly available dataset obtained from RecSys Challenge in 2015. This dataset contains a collection of sequences of click events with respective timesteps from Yoochoose website. Some of the click sessions ended with a purchase event (if so, label was set as positive, otherwise negative). This dataset reflects on reproduceability of retargeting results from this study only, as there is no publicly available prospecting dataset to the best of our knowledge.
	
\paragraph{User activity trails from Verizon Media.}
We also conducted experiments using user activity trails data from Verizon Media. This includes activities done in chronological order by a user, and the activities are derived from heterogeneous sources, e.g., Yahoo Search and Mail, reading news and other content on publisher's webpages associated with Yahoo, advertising data from Yahoo Gemini and Verizon Media DSP and data from all advertisers  (e.g., ad clicks, conversions, and site visits). The representation of an activity comprises of activity ID, time stamp, its type (e.g., search, invoice, reservation, content view, order confirmation, parcel delivery), and a raw description of the activity (e.g., the exact search query for search activities) after stripping personally identifiable information. 
To ensure legality of information used, datasets created for each advertiser strictly follow legal guidelines determined by the contract, i.e. data collected from advertiser A will never be used for any optimization task for advertiser B.

Site visits are events that are most commonly labeled as retargeting events, i.e. user who is browsing to buy a furniture item on advertiser's website will in the next several months be regarded as a retargeting user for furniture conversions for that advertiser. As mentioned in the Introduction, advertisers who focus on prospective advertising are interested in generating new converters from non-retargeting users (users who did not visit advertiser's website), however, learning to target prospecting users from the existing data is very difficult. Namely, a common theme for a majority of \textit{retail advertisers} is that a user will visit their webpage at least once before purchasing anything. Conversions in terms of whether a user visited advertisers' website for a single major retail advertiser are characterized in Table~\ref{tbl:retargeting}.
\begin{table}[h!]
	\centering
	\resizebox{0.48\textwidth}{!}{%
	\begin{tabular}{llll}
		\toprule
		Conversion & Adv. site visit & Site visit prior to conv. & Percentage \\ 
		\rowcolor[gray]{0.85}
		TRUE       & FALSE           & FALSE                     & 0.01\%     \\ 
		TRUE       & TRUE            & FALSE                     & 0.02\%     \\ 
		\rowcolor[gray]{0.85}
		TRUE       & TRUE            & TRUE                      & 99.97\%    \\ \bottomrule
	\end{tabular}}%
	\caption{Percentages of conversions with respect whether advertisers' site visit (a retargeting) event occured, and if it occurred before the conversion or not.}
	\label{tbl:retargeting}
	\vspace{-20pt}
\end{table}
Table~\ref{tbl:retargeting} clearly shows that the vast majority of conversions happen after users visited advertiser's website, thus becoming retargeting users before conversion. The goal of  DSP prospective targeting is to target users before they become retargeting users, thus bringing new users to the advertiser and boosting their sales.

\begin{figure}[ht!]
	\centering
	\includegraphics[width=0.45\textwidth]{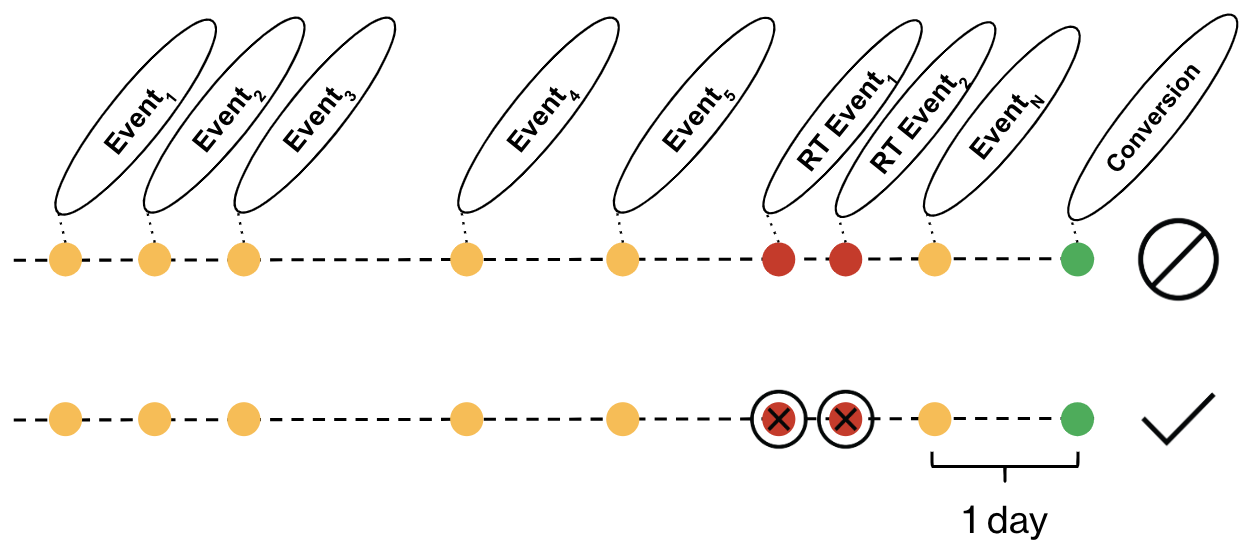}
	\caption{Visualisation of trail cutting process event before retargeting event happens.}
	\label{fig:uat_trail_cutting}
	\vspace{-10pt}
\end{figure}

Any algorithm trained on original data collected will be biased towards modeling retargeting signals only as simple rule-based modes such as predicting conversion for all users that visit advertisers' website will yield very high recall (i.e. 99.7\%). 
To prevent this from happening, we are performing a retargeting events blacklisting, as highlighted by the advertiser. The process is shown in Fig.~\ref{fig:uat_trail_cutting} and it reflects use cases where the algorithm learns to predict whether a user is going to convert the next day or not based on all signals, and not simply looking at site visits.

Dataset used in this study is collected from a single anonymized major advertiser who defined two different conversion rules and it comprises (after eligible users and events are selected and negatives downsampling is performed to maintain roughly 10\% of positives) of $788,551$ users in train and $196,830$ for test set collected over an undisclosed period longer than $90$ days.

\section{Experiments}
\label{sec:experiments}
We first describe baseline algorithms that can capture information from the sequences of events as such models reportedly outperform standard structured models. Evaluation metrics are then defined. And finally results on both public and proprietary dataset are provided and discussed.

\subsection{User Modeling Baselines}
\label{sec:baselines}
The following models are selected to either represent previously published studies or as models that are expected to fit well with the given setup.
\begin{enumerate}
    \item Recurrent Neural Network (RNN):
        A recurrent neural network with embedding layer and GRU cells to ensure fast convergence.
    \item 1-dimensional Convolutional Neural Networks (CNN):
        A 1-dimensional Convolutional neural network on top of learned event embeddings.
    \item RNN with attention layer (RNN+Attn):
        An extension of RNN model with additional attention layer used to summarize the sequence \cite{gligorijevic2018sdm}.
    \item RNN with self attention layer (RNN+SelfAttn):
        Another extension of RNN model with self-attention layer used learn higher order interactions between events before the RNN block \cite{vaswani2017attention}.
\end{enumerate}

\subsubsection{Evaluation metrics}
For assessing the quality of estimated CVR probabilities, we use the area under the ROC curve (AUC) classification performance measure, in addition to Accuracy, Precision and Recall obtained after choosing the appropriate classification threshold. 

In addition, for the proprietary data, we study the bias \cite{Baeza-Yates:1999:MIR:553876} of the predicted probabilities defined as ratio between sum of sample ($s \in S$) conversion probabilities $p(s) \in [0 .. 1]$, and sum of conversion labels $l(s) \in \{0,1\}$ as $Bias = \frac{\sum_{s \in S}p(s)}{\sum_{s \in S} l(s)}$.
% \begin{equation}
%     Bias = \sum_{s \in S} p(s)/\sum_{s \in S} l(s)
% \end{equation}. 
Unbiasedness (Bias=1) is a desirable property, as higher than 1 bias implies overly-optimistic estimates and waste of resources (bidding where there is a lower chance of conversion), and lower than 1 bias implies to overly-conservative estimates and missed opportunity (not bidding where there is a higher probability of conversion).

Finally, for the cases of multi--task learning where class disbalance becomes prominent we report area under Precision-Recall curve, as a more representative metric \cite{davis2006relationship}. 

\subsection{Experimental results}
The proposed algorithm and baseline methods are evaluated on two described datasets and the results are given below.
\subsubsection{Results on public dataset}
\label{sec:results_youchoose}
Results of the experiments on public data source are given in Table~\ref{tbl:youchoose_results}.
\begin{table}[h!]
	\resizebox{0.48\textwidth}{!}{%
	\begin{tabular}{lllllll}
		\toprule
		 & ROC AUC & PRC AUC & Accuracy & Precision & Recall \\ \midrule
		 \rowcolor[gray]{0.85}
		CNN & 0.7534 & 0.2870 & 0.6779 & 0.2087 & 0.7041 \\ 
		GRU & 0.7504 & 0.2725 & 0.6958 & 0.2142 & 0.6746 \\ 
		\rowcolor[gray]{0.85}
		GRU+SelfAttn & 0.7029 & 0.2391 & 0.6734 & 0.1907 & 0.6184 \\ 
		GRU+Attn & 0.7639 & 0.2973 & \textbf{0.6997} & \textbf{0.2195} & 0.6904 \\ 
		\rowcolor[gray]{0.85}
		DTAIN & \textbf{0.7666} & \textbf{0.3019} & 0.6943 & 0.2186 & \textbf{0.7047} \\ \bottomrule
	\end{tabular}}%
	\caption{Performance metrics on the Youchoose dataset for all algorithms.}
	\label{tbl:youchoose_results}
	\vspace{-18pt}
\end{table}
The ROC AUC and PRC AUC results show that the proposed GRU+TimeAttn model outperforms all of the baselines. The PRC AUC was reported as the ratio of positives was approximately 9\% in the dataset. Competitive results in the rest of the metrics show that the temporal information can truly help the predictive task even in datasets such as this one, expecially Given that all examples in the public dataset occur withing one hour time window. It may be surprising that adding temporal information helps, however, as discussed in the Section~\ref{sec:proposed_approach}, the temporal information has two aspects to it and thus can model initial impact of the events to the conversion thus providing additional information to the classifier. 

\subsubsection{Results on proprietary dataset - prospecting users conversion prediction}
\label{sec:results_vzm_retargeting}
\paragraph{Results on binary classification} 
In this section we conduct experiment on prediction task whether user converted for any of the conversion rules set by the advertiser. Similar, yet more prominent results are obtained on the proprietary dataset where temoral aspect plays a major role in prediction (Table~\ref{tbl:retargeting_results}).
\begin{table}[h!]
	\resizebox{0.48\textwidth}{!}{%
	\begin{tabular}{llllll}
		%&  &  &  &  &  \\ 
		\toprule
		\multicolumn{1}{l}{} & \multicolumn{1}{l|}{ROC AUC} & \multicolumn{1}{l}{Accuracy} & \multicolumn{1}{l}{Precision} & \multicolumn{1}{l}{Recall}  & \multicolumn{1}{l}{Bias} \\ \midrule
		\rowcolor[gray]{0.85}
		\multicolumn{1}{l}{CNN} & \multicolumn{1}{l}{0.8806}  & \multicolumn{1}{l}{0.8110} & \multicolumn{1}{l}{0.2457} & \multicolumn{1}{l}{0.7871} & \multicolumn{1}{l}{1.0161} \\ 
		\multicolumn{1}{l}{GRU} & \multicolumn{1}{l}{0.9018}  & \multicolumn{1}{l}{0.8520} & \multicolumn{1}{l}{0.3004} & \multicolumn{1}{l}{0.7972} &  \multicolumn{1}{l}{1.1983} \\ 
		\rowcolor[gray]{0.85}
		\multicolumn{1}{l}{GRU+Attn} & \multicolumn{1}{l}{0.8968} & \multicolumn{1}{l}{0.8438} & \multicolumn{1}{l}{0.2882} & \multicolumn{1}{l}{0.7982} &  \multicolumn{1}{l}{0.8047} \\ 
		\multicolumn{1}{l}{GRU+SelfAttn} & \multicolumn{1}{l}{0.8804} & \multicolumn{1}{l}{0.8364} & \multicolumn{1}{l}{0.2743} & \multicolumn{1}{l}{0.7756} & \multicolumn{1}{l}{0.9273} \\ 
		\rowcolor[gray]{0.85}
		\multicolumn{1}{l}{DTAIN} & \multicolumn{1}{l}{\textbf{0.9263}} & \multicolumn{1}{l}{\textbf{0.8602}} & \multicolumn{1}{l}{\textbf{0.3219}} & \multicolumn{1}{l}{\textbf{0.8537}} & \multicolumn{1}{l}{\textbf{0.9871}} \\ \bottomrule
	\end{tabular}}%
	\caption{Performance metrics on the proprietary user trails dataset for all algorithms.}
	\label{tbl:retargeting_results}
	\vspace{-20pt}
\end{table}
We can see that the DTAIN outperforms other baselines by a large margin on all metrics. The time aspect of the events is much more prominent in the proprietary dataset. Moreover, as the time window is significantly larger, the events may repeat multiple times, and time mechanism will be able to select the most important events out of the redundant ones and thus filter out the noise in the data.

\paragraph{Results on multi--task classification}
Finally we show results for the multi--task classification setup where we predict whether user will not convert for the advertiser, or for which of the two conversion rules will the user convert.
\begin{table}[h!]
	\resizebox{0.48\textwidth}{!}{%
	\begin{tabular}{llllll}
	    \toprule
		& PRC AUC & Accuracy & Precision & Recall & Bias  \\
		\midrule
		Task 0    \\
		\midrule
		\rowcolor[gray]{0.85}
		CNN & 0.9880 & 0.8139 & 0.9810 & 0.8153 & 1.0069 \\
		GRU & 0.9896 & 0.8544 & 0.9821 & 0.8588 & 1.0030 \\
		\rowcolor[gray]{0.85}
		GRU+Attn & 0.9907 & 0.8511 & 0.9837 & 0.8537 & 0.9933 \\
		GRU+SelfAttn & 0.9877 & 0.8456 & 0.9795 & 0.8515 & 0.9941 \\
		\rowcolor[gray]{0.85}
		DTAIN & \textbf{0.9926} & \textbf{0.8613} & \textbf{0.9876} & \textbf{0.8614} & \textbf{0.9982} \\
		\midrule
		Task 1\\
		\midrule
		\rowcolor[gray]{0.85}
		CNN & 0.2523 & 0.9602 & 0.3161 & 0.2506 & 0.8836 \\
		GRU & 0.2711 & 0.9629 & 0.3635 & 0.2720 & \textbf{0.9715} \\
		\rowcolor[gray]{0.85}
		GRU+Attn & \textbf{0.3013} & 0.9630 & 0.3788 & \textbf{0.3139} & 1.1163 \\
		GRU+SelfAttn & 0.2452 & 0.9606 & 0.3277 & 0.2648 & 1.0645 \\
		\rowcolor[gray]{0.85}
		DTAIN & \textit{0.2880} & \textbf{0.9652} & \textbf{0.4000} & 0.2539 & 1.0680 \\
		\midrule
		Task 2 \\
		\midrule
		\rowcolor[gray]{0.85}
		CNN & 0.2495 & 0.9584 & 0.3287 & 0.2419 & 0.7588 \\
		GRU & 0.2567 & 0.9597 & 0.3485 & 0.2464 & 0.8849 \\
		\rowcolor[gray]{0.85}
		GRU+Attn & 0.2374 & 0.9584 & 0.3355 & \textbf{0.2582} & 1.0453 \\
		GRU+SelfAttn & 0.2081 & 0.9587 & 0.3081 & 0.1951 & \textbf{0.9887} \\
		\rowcolor[gray]{0.85}
		DTAIN & \textbf{0.2776} & \textbf{0.9633} & \textbf{0.4083} & 0.2348 & 0.9460 \\
		\bottomrule
	\end{tabular}}%
	\caption{Performance metrics on the proprietary user trails dataset for different tasks.}
\end{table}
\vspace{-20pt}
As the positives and negatives rate becomes very disbalanced when binary task is split to multi-task classification tasks, we report PRC AUC metric \cite{davis2006relationship}. 
The DTAIN shows the best performance on majority of metrics across the three tasks always having the top Accuracy and Precision metrics. Overall evaluation shows that the DTAIN model is the best among the chosen baselines once again. The DTAIN model was prominently the best approach for the Task 0 (prediction if the user is not going to convert) which is very important for the bidding system to know if it should bid for a user or not.

\subsubsection{Attention analysis and interpretation}
\begin{figure}[htp]
  \centering
  \label{figur}\caption{Heat maps of events attentions scores for 100 randomly sampled converters}

  \subfloat[GRU+Attn attention]{\label{fig:gru_atttn}\includegraphics[width=40mm]{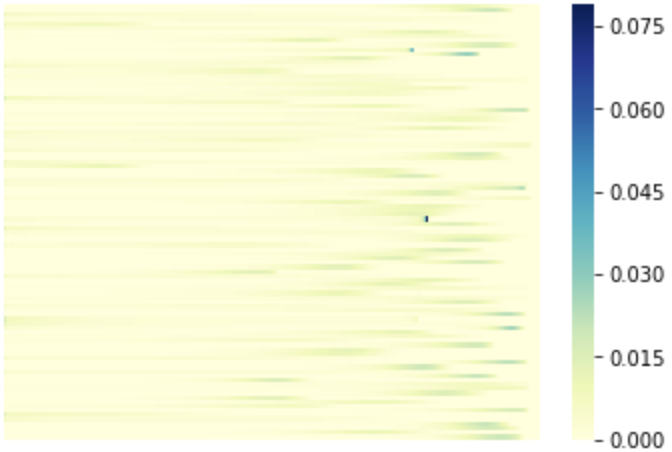}}
  \subfloat[DTAIN attention]{\label{fig:dtain_attn}\includegraphics[width=40mm]{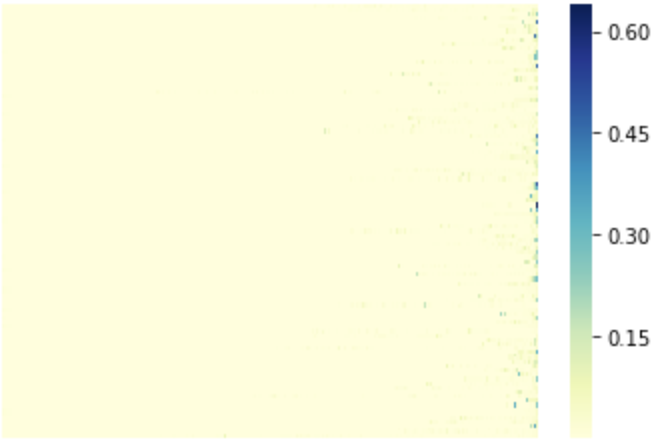}}
  \\
  \subfloat[DTAIN $\theta_{e_i}$]{\label{fig:dtain_theta}\includegraphics[width=40mm]{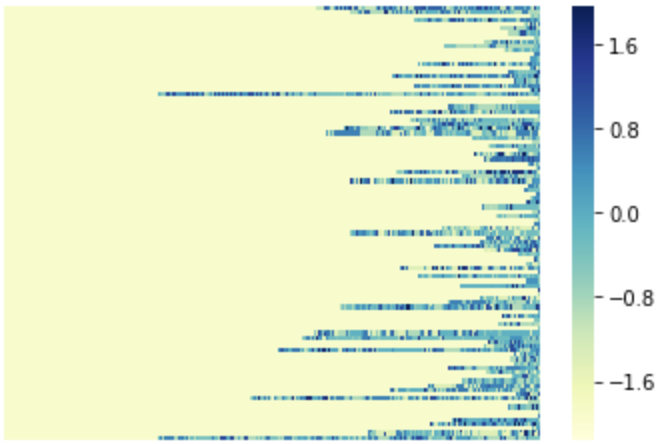}}
  \subfloat[DTAIN $\mu_{e_i}$]{\label{fig:dtain_mu}\includegraphics[width=40mm]{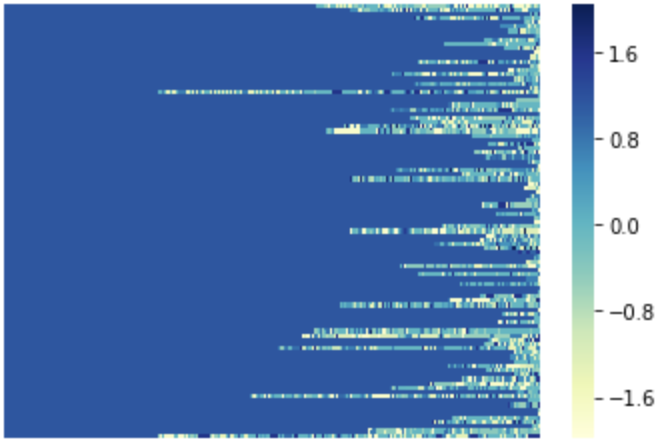}}
\end{figure}
\vspace{-10pt}
To tap into the explainability of the models we randomly selected a hundred converters and analyzed attentions of their events. We compare DTAIN model primarily against the GRU+Attn model, which has shown properties of explainability in the past \cite{gligorijevic2018sdm}.  
From Fig.~\ref{fig:gru_atttn} it can be seen that GRU+Attn model assigns attentions across the users trails, highlighting not only events that happened close to conversion which is a desirable property for prospective advertising. 
The DTAIN model has a slightly different mechanism of attention as time plays a major role in allowing information from different signals to be passed through the network. As discussed in Section~\ref{sec:methodoogy} key parameters $\theta_{e_i}$ and $\mu_{e_i}$ have interesting interpetability properties. To show this we plot scores of both the key parameters in Fig.~\ref{fig:dtain_theta} and ~\ref{fig:dtain_mu} respectively, and of attentions from the attention block in Fig.~\ref{fig:dtain_attn}. Interestingly, we can see that there are plenty of high positive values of $\theta_{e_i}$ and high negative values of $\mu_{e_i}$ further away from the end of sequences, in addition to the expected ones closer to the end of it. This means that the DTAIN is capturing long term as well as short term patterns and controls which events signals fully pass through the rest of the network. Moreover, interesting pattern shows when we observe the attention scores, which look very different from the ones from the GRU+Attn model as they are focused towards the end of the sequence. It is important to notice that less relevant events have been already filtered by the time-aware mechanism before being passed into the GRU layer and overall signals of sequences are then summarized in last few vectors in the GRU layer output, which was not possible in the other model. 

These interesting findings allow us to use the attention scores for explainabilty to the advertisers by providing them insights into both long- and short-term patterns and important events that they can further use to improve their creatives and advertising strategies.

\section{Conclusions and Future Work}
\label{sec:conclusions}
In this study we proposed a sequence based approach for modeling conversion prediction based on users' activity trails that leverages both the sequence and temporal information of heterogeneous events collected from many data sources. We proposed a new way to model temporal information for conversion prediction that preserves ability of interpretation, and finally we showed that the DTAIN mode outperforms baselines that represent state-of-the art on both public and proprietary datasets. However, as the data is collected from many data sources, and different events may repeat often or periodically there it is still significant noise that the algorithms need to address and developing novel techniques to address these concerns will be the next steps in developing new solutions.

\balance
\bibliographystyle{abbrv}
\bibliography{sigproc}

\end{document}